\documentclass{article}

\usepackage[final]{nips_2017}
\usepackage{graphicx}
\usepackage{comment}
\usepackage{amsmath}
\usepackage{booktabs}

\title{Ubenwa: Cry-based Diagnosis of Birth Asphyxia}

\author{Charles C. Onu$^{1,2}$,
Innocent Udeogu$^{2}$,
Eyenimi Ndiomu$^{2}$, 
Urbain Kengni$^{2}$, 
Doina Precup$^{1}$, 
\\ Guilherme M. Sant'anna$^{3}$, 
Edward Alikor$^{4}$ and 
Peace Opara$^{4}$
\thanks{
$^{1}$ C. C. Onu and D. Precup are with the School of Computer Science, McGill University, Montreal, Canada. (e-mail: {charles.onu@mail.mcgill.ca}).
$^{2}$ C. C. Onu, I. Udeogu, E. Ndiomu and U. Kengni are with Ubenwa Intelligence Solutions Inc.
$^{3}$ G. M. Sant'anna is with the Division of Neonatology, McGill University Health Centre. 
$^{4}$ E. Alikor and P. Opara are with the University of Port Harcourt Teaching Hospital (UPTH), Nigeria}
}

\begin{document}

\maketitle
\thispagestyle{empty}
\pagestyle{empty}

\begin{abstract}

Every year, 3 million newborns die within the first month of life. Birth asphyxia and other breathing-related conditions are a leading cause of mortality during the neonatal phase. Current diagnostic methods are too sophisticated in terms of equipment, required expertise, and general logistics. Consequently, early detection of asphyxia in newborns is very difficult in many parts of the world, especially in resource-poor settings. We are developing a machine learning system, dubbed Ubenwa, which enables diagnosis of asphyxia through automated analysis of the infant cry. Deployed via smartphone and wearable technology, Ubenwa will drastically reduce the time, cost and skill required to make accurate and potentially life-saving diagnoses.

\end{abstract}

\section{Problem Statement}
In 2016, nearly 3 million babies died within 28 days of coming into the world - the neonatal period [1]. Birth Asphyxia has been identified by the World Health Organisation (WHO) and other public health organisations [1,2,3] as one of the top 3 causes of newborn mortality globally. It also results to severe, life-long disabilities (such as cerebral palsy, deafness, and intellectual difficulty) in over 1 million infants, annually [4].

\begin{figure}[htbp]
   \centering
   \includegraphics[width=.35\textwidth]{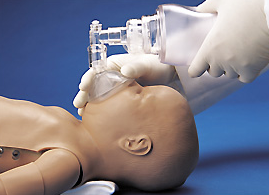}
   \caption{Example of immediate manual resuscitation recommended for a severely asphyxiating infant}
   \label{fig:manualresus}
 \end{figure}

More than half of under-5 child deaths are due to diseases that are preventable and treatable through simple, affordable interventions [1] (example in Fig \ref{fig:manualresus}). Unfortunately, the opportunity for early detection is limited in many resource-poor settings due to the logistics and cost of existing diagnostic system. Consequently, breathing conditions like asphyxia are generally detected only when the visual symptoms (such as pale/bluish limbs) have emerged, at which point severe neurological damage may have already occurred.


Confirmatory clinical diagnosis of asphyxia involves analysis of an arterial blood sample of the infant to measure blood gases, pH, oxygen saturation and electrolytes, using a blood gas analyser [5]. This information combined with the APGAR score - a standard physical assessment of the newborns based on 5 parameters - gives conclusive evaluation of the presence and/or severity of asphyxia. Whereas blood gas analysis is a routine procedure for newborns in developed countries, in many developing regions of the world it is not. Thus, a high proportion of morbidity and mortality resulting from birth asphyxia occurs in low and middle income countries [1].

There is need for diagnostic methods for birth asphyxia which lend themselves to early detection. Additionally, if such method is cheap, and easy to use, community health workers, midwives, nurses and even parents can use it to make life-saving diagnosis of asphyxia possible in nearly every corner of the world.

\section{Approach}

Previous studies have hypothesised that breathing difficulty resulting from asphyxia alters the patterns in the cry waves of affected infants, largely attributed to the fact that speech and breathing are controlled by the same underlying physiologic process [6,7]. Utilising the infant cry to diagnose asphyxia presents a unique opportunity for the development of a more accessible diagnostic tool.

In this work, we demonstrated via a retrospective study that the infant cry provides rich source of information about the physiological state of a newborn. We further developed a prototype mobile app which could be used in resource-poor settings to reliably detect birth asphyxia.

\begin{figure*}[!b]
   \centering
   \includegraphics[width=1\textwidth]{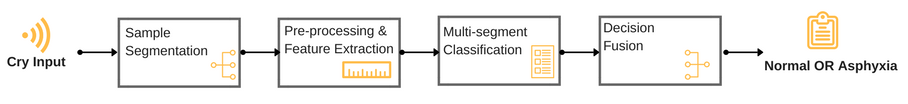}
   \caption{A high-level view of the Ubenwa diagnostic process}
   \label{fig:blockdiag}
 \end{figure*}

\subsection{Mel Frequency Cepstral Coefficients (MFCC) with Support Vector Machines (SVM)}
We leveraged techniques from automatic speech recognition [8]. We combined features extracted as coefficients of the mel frequency spectrum with a support vector machine classifier. MFCCs are widely used in automatic speech recognition problems as they provide a representation of audio signals that closely mimic the human auditory system [10]. SVMs are powerful classifiers that can learn complex, non-linear decision boundaries. Compared to other non-linear classifiers like neural networks, SVMs are designed to work effectively with limited examples and high-dimensional data [11], as is the case in our problem.

Fig. \ref{fig:blockdiag} provides a overview of our diagnostic system. Samples are first broken down into time segments. Each segment undergoes pre-processing (e.g., removing leading and trailing blanks) and feature extraction (as MFCC). The MFCCs are fed as input to the SVM classifier. A sample is classified as normal or \textit{asphyxia}, if the majority of its segments were classified as such.

\section{Results}
\label{section:results}

We obtained the Baby Chillanto Database courtesy of the National Institute of Astrophysics and Optical Electronics, CONACYT, Mexico [7]. The database contains cries of 69 normal, asphyxiating and deaf infants (deafness is one of the most common disabilities resulting from asphyxia). Of interest to our work were the cry samples of normal and asphyxiating newborns. These were further synthesised into 1389 samples.

80\% of the data was used to train and validate the algorithms while 20\% was kept aside as a test set. Results on this test set showed sensitivity (accuracy in detecting asphyxiating infants) and specificity (accuracy in detecting normal infants) of 85\% and 89\%, respectively [11].

\subsection{Prototype}
\begin{figure*}[bp]
   \centering
   \includegraphics[width=.7\textwidth]{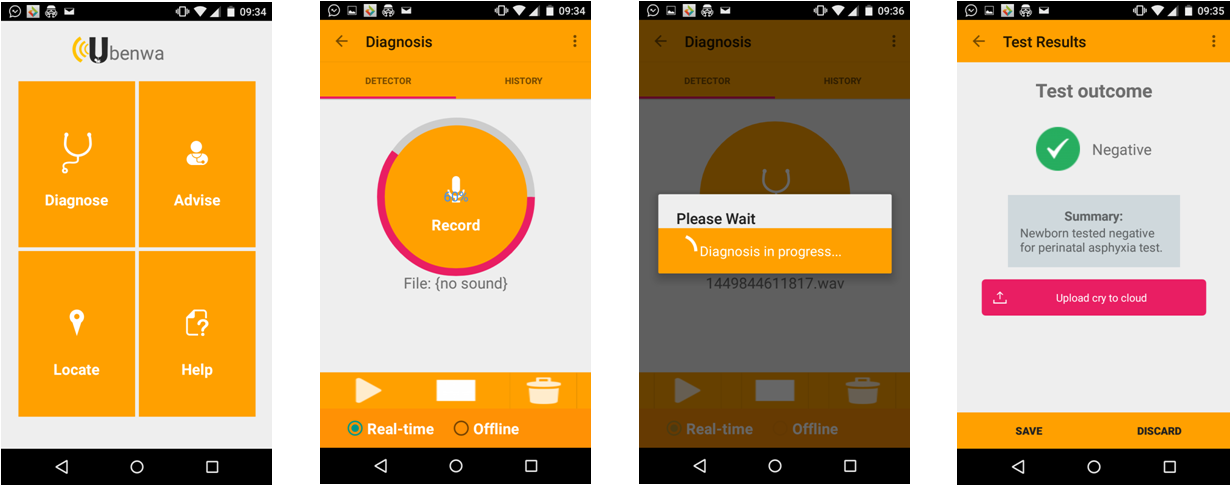}
   \caption{Actual screenshots of the Ubenwa mobile app}
   \label{fig:ubenwademo}
 \end{figure*}

We acknowledge that machine learning is only a part of the solution. For deployment, we built a mobile application into which we have incorporated our current model. We have dubbed it {\em Ubenwa} which means "cry of a baby" in Igbo language of Nigeria. Ubenwa has allowed us to think deeply about our ultimate objective of deploying more accessible diagnostic tool. It has also given us opportunity to receive feedback from clinicians and other caregivers. Screenshots of the Ubenwa are shown in Fig. \ref{fig:ubenwademo}.

The use of the infant cry as input for diagnosis of asphyxia presents significant economic, social and clinical benefits. Concretely, compared to the current method using a blood gas analyser, Ubenwa is:

\begin{enumerate}
\item non-invasive (requiring only cry rather than blood), 
\item low-cost (only as expensive as the cost of a phone), 
\item requires little or no skill to operate, and 
\item delivers results much quicker (under 20 seconds).
\end{enumerate}
Using Ubenwa, birth attendants, parents and other care-givers can quickly detect asphyxia in newborns, and promptly refer them for potentially life-saving treatment.



\subsection{MFCC/SVM vs Deep learning}
Until the recent surge in deep learning methodologies, the combination of MFCC and SVMs gave some of the best performances on for speech recognition tasks [12]. Today, recurrent neural network (RNN)-based models such as long short-term memory (LSTM) networks have led to tremendous gains on many speech and language problems [13 - 15].  We applied MFCC/SVM in our pilot work as it provided 2 critical advantages over a deep learning approach including feasibility of training under small number of examples and portability of model within a mobile application. As we acquire more data, and seeing that  frameworks for deployment of deep models to mobile platforms are beginning to mature[16], we will develop deep neural networks which could potentially raise performance on this task of asphyxia detection to levels of clinical utility.


\section{Next Steps}
\label{section:nextsteps}
Our immediate focus is to carry out data collection over a one year period (starting January 2018) at 2 selected sites in Canada and Nigeria. It is necessary to obtain more cry samples (especially of the pathological case) in order to train our algorithm to do better at correctly classifying new subjects.

\subsection{Data acquisition and product validation plans}
The data acquisition will serve as an opportunity to validate our current system in a real world scenario. In particular, we plan to carry out data acquisition at 2 hospitals: University of Port Harcourt Teaching Hospital (UPTH), Port Harcourt, Nigeria and McGill University Health Centre (MUHC), Montreal, Canada. We have applied for and obtained approval from the institution ethics board at UPTH. We are currently working on the same at the MUHC.

\subsection{Optimisations}
We will continue to work on several lines of optimisations for our algorithm, namely: training for robustness to noise in the environment, finding the shortest possible record length for which we can make accurate diagnosis, optimising (math) operations to ensure that algorithm requires minimal memory and computation on mobile devices,  and others.


\addtolength{\textheight}{-12cm}   






\section*{References}
[1] World Health Organisation, “Children: Reducing mortality,” Fact Sheet,
http://www.who.int/mediacentre/factsheets/fs178/en/, 2016.

[2] United Nations, “We can end poverty: Millenium development goals and
beyond 2015,” 2013.

[3] J. Lawn and K. Kerber, “Opportunities for africas newborns: practical
data policy and programmatic support for newborn care in africa.” 2006.

[4] World Health Organization, “The world health report,” 2005.

[5] J. Low, “Intrapartum fetal asphyxia: Definition, diagnosis, and classification,” vol. 176, pp. 957–9, 06 1997.

[6] K. Michelsson, P. Sirvi, and O. Wasz-Hckert, “Pain cry in full-term
asphyxiated newborn infants correlated with late findings,” vol. 66, pp.
611–6, 10 1977.

[7] O. F. Reyes-Galaviz and C. A. Reyes-Garcia, “A system for the
processing of infant cry to recognize pathologies in recently born babies
with neural networks,” in 9th Conference Speech and Computer, 2004.

[8] A. Waibel and K.-F. Lee, Readings in speech recognition. Morgan
Kaufmann, 1990.

[9] M. R. Hasan, M. Jamil, M. G. Rabbani, and M. S. Rahman, “Speaker
identification using mel frequency cepstral coefficients,” variations,
vol. 1, no. 4, 2004.

[10] C. Cortes and V. Vapnik, “Support-vector networks,” Machine learning,
vol. 20, no. 3, pp. 273–297, 1995.

[11] C. C. Onu, “Harnessing infant cry for swift, cost-effective diagnosis
of perinatal asphyxia in low-resource settings,” in Humanitarian Technology
Conference-(IHTC), 2014 IEEE Canada International. IEEE,
2014, pp. 1–4.

[12] B. Panda, D. Padhi, K. Dash, "Use of SVM Classifier \& MFCC in Speech Emotion Recognition System", IJARCSSE, vol. 2, no. 3, March 2012.

[13] Y. LeCun, Y. Bengio, and G. Hinton, “Deep learning,” Nature, vol. 521,
no. 7553, pp. 436–444, 2015.

[14] I. Sutskever, O. Vinyals, and Q. V. Le, “Sequence to Sequence Learning with Neural Networks,” Sept.2014.

[15] A. Graves, A. Mohamed, and G. Hinton, “Speech recognition with deep recurrent neural networks,” in 2013 IEEE International Conference on Acoustics, Speech and Signal Processing, 2013, pp. 6645–6649.

[16] M. Abadi, P. Barham, J. Chen, Z. Chen, A. Davis, J. Dean, M. Devin,
S. Ghemawat, G. Irving, M. Isard, M. Kudlur, J. Levenberg, R. Monga,
S. Moore, D. G. Murray, B. Steiner, P. A. Tucker, V. Vasudevan,
P. Warden, M. Wicke, Y. Yu, and X. Zhang, “Tensorflow: A system
for large-scale machine learning,” CoRR, vol. abs/1605.08695, 2016.
[Online]. Available: http://arxiv.org/abs/1605.08695

\end{document}